\documentclass[10pt,letterpaper]{article}
\usepackage[T1]{fontenc}
\usepackage[utf8]{inputenc}
\usepackage[margin=0.875in]{geometry}
\usepackage{amsmath,amssymb,graphicx,booktabs,array}
\usepackage[numbers,super,sort&compress]{natbib}
\usepackage[hidelinks]{hyperref}
\usepackage{caption}
\newcommand{\R}{\mathbb{R}}
\newcommand{\diag}{\operatorname{diag}}
\newcommand{\perc}{\operatorname{perc}_{95}}
\title{\bfseries Parameter-Efficient Fine-Tuning of Foundation Models\\for Liver Tumor Segmentation in CT}
\author{\normalsize Ramtin Mojtahedi\textsuperscript{a,b}, Mohammad Hamghalam\textsuperscript{a,f}, Jacob J. Peoples\textsuperscript{c}, Richard K. G. Do\textsuperscript{c},\\
\normalsize and Amber L. Simpson\textsuperscript{d,e}}
\date{}
\begin{document}
\maketitle
\vspace{-1.5em}
\begin{center}\small
\textsuperscript{a}School of Computing, Queen's University, Kingston, ON, Canada\\
\textsuperscript{b}Toronto General Research Institute, University Health Network, Toronto, ON, Canada\\
\textsuperscript{c}Department of Radiology, Memorial Sloan Kettering Cancer Center, New York, NY, USA\\
\textsuperscript{d}Department of Radiology and Diagnostic Imaging, University of Alberta, Edmonton, AB, Canada\\
\textsuperscript{e}Alberta Machine Intelligence Institute, Edmonton, AB, Canada\\
\textsuperscript{f}Department of Electrical Engineering, Qa.C., Islamic Azad University, Qazvin, Iran
\end{center}
\begin{center}\footnotesize
Author manuscript. Published as: R. Mojtahedi et al., ``Parameter-efficient fine-tuning of foundation models for liver tumor segmentation in CT,'' \emph{Medical Imaging 2026: Computer-Aided Diagnosis}, Proc. SPIE \textbf{13926}, 1392612, pp. 260--268 (2026). \url{https://doi.org/10.1117/12.3087835}.
\end{center}
\begin{abstract}
State-of-the-art interactive segmentation with foundation models remains constrained by compute and the need for annotations. We evaluated parameter-efficient fine-tuning (PEFT) of the Segment Anything Model (SAM) by inserting baseline adapters, Low-Rank Adaptation (LoRA), 4-bit Quantized Low-Rank Adaptation (QLoRA), and a convolutional adapter (Conv-Adapter), as well as our proposed Directional Spectral Top-K adapter (DiSCo), training only the adapters while keeping the SAM backbone frozen. DiSCo performs singular value decomposition of row-normalized weights to obtain spectral bases, then learns rank-gated spectral coefficients, per-output magnitude offsets, and a spectral gain; it supports optional Top-K rank selection at inference while keeping only 0.14 M parameters trainable. We benchmarked five prompting regimes, none, single-point, multi-point, and bounding boxes at an intersection over union (IoU) of 0.50 and 0.75, applied to abdominal CT scans of colorectal liver metastases, and reported segmentation metrics (Dice and the 95th-percentile Hausdorff distance (HD95)) and compute metrics (trainable parameters, latency, memory, throughput). Conv-Adapter and LoRA achieved the highest accuracy (overall Dice 0.793 and 0.792; single-point 0.795 and 0.792; HD95 32 mm). QLoRA was close (overall 0.766; single-point 0.768; HD95 36.41 mm) while offering the most favorable compute profile (0.91 M trainable parameters, 120 ms latency, 4.9 GB peak memory). DiSCo maximized parameter efficiency, achieving the highest Dice per million trainable parameters (4.66), showing an accuracy--efficiency trade-off (overall Dice 0.653; single-point 0.698; HD95 49.53 mm). These results indicate that PEFT on foundation models enables more accurate liver tumor segmentation with reduced adaptation costs, supporting rapid scanner-specific tuning, point- or box-prompt workflows, and broader deployability for preoperative volumetrics when compute and labeled data are limited. The code supporting this study is available at: \url{https://github.com/Ramtin-Mojtahedi/PEFT-SAM-Liver-CT}.
\end{abstract}
\begingroup\renewcommand{\thefootnote}{}\footnotetext{Further author information: (Send correspondence to R.M.)\\R.M.: E-mail: \href{mailto:ramtin.mojtahedi@queensu.ca}{ramtin.mojtahedi@queensu.ca}\\R.M.: Current affiliation: Toronto General Research Institute, University Health Network, Toronto, ON, Canada}\endgroup

\section{Introduction}
Colorectal cancer is the third most common cancer globally and is one of the leading causes of death from cancer\cite{ref1} due to metastatic disease, most commonly in the liver. Contrast-enhanced computed tomography (CT) images are used to assess liver lesions and surrounding hepatic parenchyma to support surgical planning, longitudinal follow-up, and quantitative imaging.\cite{ref2,ref3,ref4,ref5} The labor-intensive and time-consuming nature of manual tumor delineation leads to inconsistencies between different readers. Therefore, the development of both automated and interactive methods to improve the reproducibility of delineations and reduce the workload of clinicians has been explored.\cite{ref6} Automated deep learning methods, including convolutional architectures, transformer-based models, and contrastive self-supervised learning have improved practitioners' ability to accurately delineate and classify tumors.\cite{ref7,ref8,ref9,ref10,ref11,ref12,ref13,ref14} Interactive segmentation further reduces annotation burden by incorporating user inputs, such as clicks, to refine segmentation masks, aligning with traditional clinical correction workflows.\cite{ref15,ref16}

Promptable foundation models including the Segment Anything Model (SAM) have improved the generalizability of models with the use of conditional segmentation and prompts such as points, boxes and text.\cite{ref17} Because medical images are quite different from natural images, using SAM directly for a zero-shot prediction may not always be reliable, thus leading to the development of MedSAM and SAM-Med2D and the use of adapter-based strategies designed to retain the pre-trained backbone but improve fidelity within the medical domain.\cite{ref18,ref19,ref20,ref21,ref22} Fully fine-tuning large transformer backbones is computationally expensive and often requires substantial labeled training data, which can be difficult to obtain in clinical practice. Parameter-efficient fine-tuning (PEFT) addresses these challenges by training lightweight modules or structured low-rank updates (e.g., adapters, LoRA, QLoRA) while keeping the backbone frozen, reducing the memory footprint and training cost.\cite{ref23,ref24,ref25}

In this study, we evaluated PEFT for adapting SAM to liver tumor segmentation in CT scans of patients with colorectal liver metastases (CRLM). We compared convolutional adapters (Conv-Adapter), Low-Rank Adaptation (LoRA), 4-bit quantized LoRA (QLoRA), and our proposed Directional Spectral Top-K adapter (DiSCo), which derives update directions from spectral (SVD) bases of row-normalized weights and learns compact rank-gated coefficients with per-output magnitude offsets and an optional inference-time Top-K constraint. We used multiple prompting regimes (no prompt, point prompts, and bounding boxes) and reported segmentation quality (Dice, HD95) as well as the deployment-relevant compute metrics (trainable parameters, latency, throughput, and peak memory) with the intention of evaluating trade-offs between efficiency and accuracy across various PEFT techniques.

Our main contributions are:
\begin{itemize}
\item We propose DiSCo, a spectral PEFT adaptation with a rank-gating feature and an optional inference-time Top-K constraint to control the effective adaptation capacity under strict parameter limitations.
\item We evaluate DiSCo against Conv-Adapter, LoRA, and QLoRA for adapting SAM to CRLM CT while keeping the SAM backbone frozen, enabling a direct comparison of accuracy--efficiency trade-offs across various PEFT techniques.
\item We evaluate prompt-driven regimes (no prompt, single- and multi-point prompts, and bounding boxes) and report best-regime segmentation metrics (Dice, HD95) and compute-efficiency metrics (latency, memory, throughput, and trainable-parameter counts) relevant to deployment.
\end{itemize}

\section{Methods}
The subsequent subsections outline the dataset description, preprocessing steps, pipeline design, model implementation, and training methodology.

\subsection{Data Description and Preprocessing}
We utilized a cohort of 447 contrast-enhanced abdominal CT scans of CRLM patients from Memorial Sloan Kettering Cancer Center. Tumor segmentation masks and corresponding images underwent human quality review. Image resolution was heterogeneous: in-plane voxel spacing typically ranged from 0.70 to 0.98 mm, and axial slice thickness was most commonly 1.5 or 5.0 mm (median 5.0 mm; IQR 2.5--5.0 mm; range 0.8--7.5 mm). Preprocessing comprised robust image--label pairing with z-alignment; liver-mask sealing and tight liver bounding-box cropping; Hounsfield-unit clipping\cite{ref26} to $[-150,250]$ with linear scaling; per-slice binary hole filling; and 2D export at $1024\times1024$ pixels to produce PNG images and corresponding tumor-mask labels.

Liver cropping creates a more focused search area for prompt-specific segmentation, resulting in reduced background noise from whatever is in the scene, while Hounsfield Unit clipping and scaling create a consistent intensity scale across all scans that matches pretrained visual networks. The export of each 2D ($1024\times1024$) image matches the general guidelines for SAM preprocessing, where inputs are resized to a constant size prior to running inference.\cite{ref17}

\begin{figure}[tb]
\centering
\includegraphics[width=0.70\linewidth]{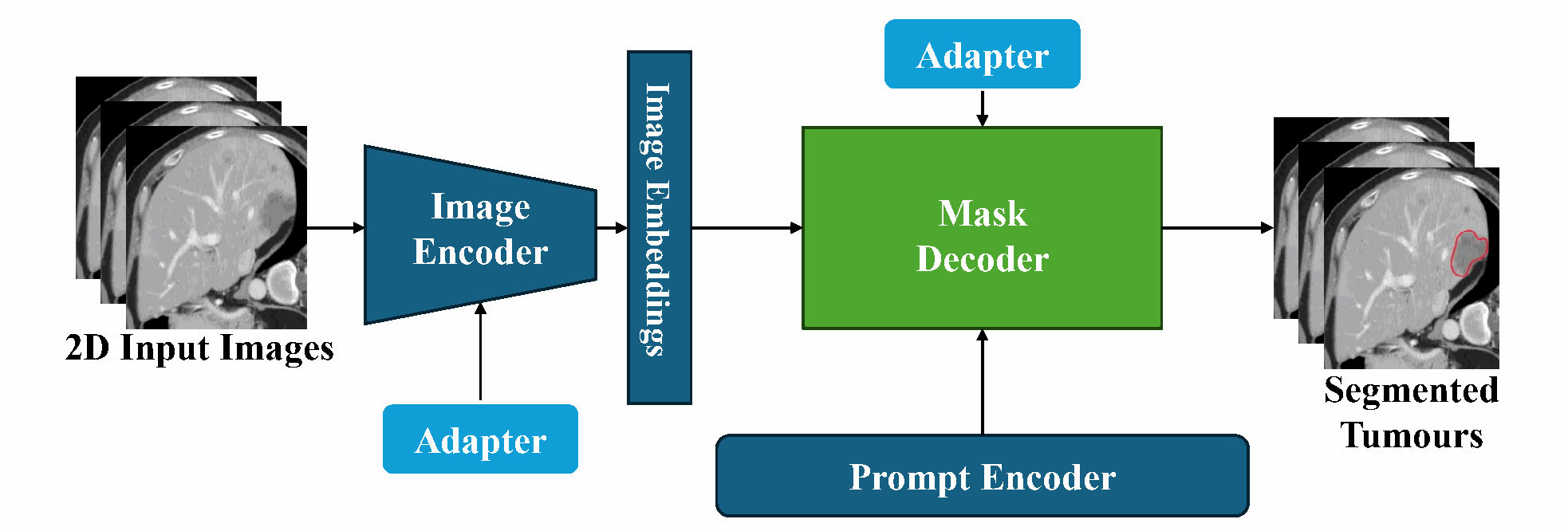}
\caption{PEFT on SAM for liver tumor segmentation in CRLM CT: we froze SAM encoders, trained only Conv-Adapter, LoRA, QLoRA, and SVD-based DiSCo adapters in the encoder or decoder while the mask decoder fused point or box prompts with image embeddings, and we evaluated multiple prompts using Dice, HD95, and compute metrics.}
\label{fig:pipeline}
\end{figure}

\subsection{Study Pipeline}
We studied parameter-efficient fine-tuning (PEFT) of SAM for liver and tumor segmentation in CT for patients with CRLM. The SAM backbone (image encoder, prompt encoder, mask decoder) was frozen, and compact adapters were inserted into selected image-encoder and mask-decoder blocks (Fig.~\ref{fig:pipeline}). We benchmarked four adapters, Conv-Adapter, LoRA, QLoRA, and our proposed DiSCo, training only adapter parameters, which yielded large reductions in trainable weights (DiSCo $\approx0.14$ M; QLoRA $\approx0.91$ M) while preserving pretrained capacity. We evaluated five prompt regimes: no prompt, a single positive point, three positive points, and bounding boxes with target IoU of 0.50 and 0.75;\cite{ref27} points and boxes were encoded by the frozen prompt encoder and injected as tokens that the mask decoder fuses with image embeddings. Point and box prompts were simulated from the ground truth tumor masks (positive points sampled within the mask and boxes derived from the tight mask bounding box) to emulate user inputs. LoRA and QLoRA were placed on selected linear projections, Conv-Adapters were used as narrow convolutional bottlenecks, and DiSCo derived spectral bases via singular value decomposition (SVD) of row-normalized reference weights and learns rank-gated spectral coefficients with per-output magnitude offsets and a spectral gain; at inference, an optional Top-K keeps only the K largest-magnitude spectral components. We reported Dice\cite{ref28} and HD95\cite{ref29} for accuracy, and trainable-parameter counts, latency, peak memory, throughput, and FLOPs\cite{ref30,ref31,ref32} for compute efficiency. For fair comparison, we used a fixed patient-level 80/20 train--validation split, a shared optimization budget, and selected checkpoints by validation Dice.

Prompt-driven evaluation aligns with interactive clinical correction as point prompts emulate sparse clicks placed inside (positive) or outside (negative) a tumor, and box prompts emulate a coarse region of interest (ROI) selection. In SAM-style pipelines, prompts are encoded as tokens by the prompt encoder and fused with image embeddings in the mask decoder, allowing a single backbone to support multiple prompt modes.\cite{ref17} Prior medical adaptations of SAM similarly evaluate click- and box-based prompting to quantify how much user prompting is required to reach clinically acceptable masks.\cite{ref18,ref19}

In order to remove adapter design influences, all of the adapter variants were trained using identical training and validation data splits and selection criterion. All evaluation metrics for the adapters were calculated at the same inference configuration; therefore, any differences along the accuracy-efficiency spectrum reflected the adapter's parameter settings, and not disparate evaluation protocols.

\subsection{Model Implementation}
\paragraph{Backbone and Scope:} We used the original Segment Anything Model (SAM) ViT-B backbone\cite{ref17} and implemented our experiments using the Medical-SAM-Adapter (Med-SA) framework.\cite{ref20} We trained only the inserted adapter parameters while keeping the SAM backbone weights frozen.\cite{ref33}

\paragraph{PEFT Rationale:} Freezing the backbone of the foundation model preserves the pretrained representation capability but allows for efficient adaptation to new domains with only a small number of parameters that can be trained. The use of frozen layers in the foundation model aligns with existing PEFT strategies for reducing the cost to store and train per task by adding lightweight modules (Adapters) or structured updates (e.g. low-rank factors) between frozen layers.\cite{ref23,ref24,ref34}

\paragraph{LoRA:} For a frozen weight $W\in\R^{d\times k}$, we inject a trainable low-rank residual as in (\ref{eq:lora}):
\begin{equation}
W'=W+\Delta W,\qquad \Delta W=\frac{\alpha}{r}AB,\quad A\in\R^{d\times r},\quad B\in\R^{r\times k},\quad r\ll\min(d,k).
\label{eq:lora}
\end{equation}
Here, $W$ is the frozen base weight and $W'$ the adapted weight; $\Delta W$ is the trainable residual; $A$ and $B$ are the LoRA factors; $d$ and $k$ are the layer's output or input dimensions; $r$ is the rank controlling capacity; and $\alpha$ scales the update as $\alpha/r$. This reduces trainable parameters from $dk$ to $r(d+k)$ while preserving quality.\cite{ref24} We integrated LoRA at 48 encoder linear sites, and the decoder and prompt encoder were unchanged.

From a practical perspective, the LoRA residual is applied without explicitly forming $\Delta W$ by computing $(AB)x=A(Bx)$ in the forward pass, which is efficient when $r$ is small. In transformer blocks, LoRA is commonly inserted into attention projections or MLP projections to provide controllable adaptation capacity with minimal memory overhead.\cite{ref24}

\paragraph{QLoRA (4-bit quantization LoRA):} Considering $W\in\R^{d\times k}$ as the Linear weight matrix, we partition $W$ into non-overlapping blocks of size $b$ and quantize each block with a 4-bit quantizer. QLoRA is integrated at 48 encoder linear sites and on top of the frozen quantized base, we train LoRA as described in (\ref{eq:qlora}):
\begin{equation}
W'=\widetilde{W}+\frac{\alpha}{r}AB,
\label{eq:qlora}
\end{equation}
where $A\in\R^{d\times r}$ and $B\in\R^{r\times k}$ are the LoRA factors, $r$ is the LoRA rank, $\alpha$ is the LoRA scaling, $b$ is the quantization block size, and $q$ specifies the 4-bit quantizer.\cite{ref25}

QLoRA reduces memory footprint by storing the base weights $\widetilde{W}$ in 4-bit precision while retaining the low-rank adapter weights trainable. This allows PEFT to be accomplished even with limited hardware resources, as the backbone storage cost can be separated from the adaptation ability of the low-rank adapter. This is most useful for large transformer backbone models that cannot be fully fine-tuned because of the prohibitive storage requirement.\cite{ref25}

\paragraph{Conv-Adapter:} We adapt frozen convolutional weights $W$ with a low-rank bottleneck and depthwise mixing; for $y=\operatorname{Conv}(x;W)$:
\begin{equation}
y'=y\odot\left(1+\frac{\delta_s}{s_0+\varepsilon}\right)+\frac{\alpha}{r}\tanh(\eta)\,B\bigl(\operatorname{DW}(A(y))\bigr),\qquad s_0=\lVert W_{i:}\rVert_2.
\label{eq:conv}
\end{equation}
Here $\delta_s$ = learnable per-channel offset, $s_0$ = per-channel norm of $W$ (with small stabilizer $\varepsilon$), $A$, $B$ = $1\times1$ convs forming a rank-$r$ bottleneck, DW = depthwise conv, $\alpha$ = adapter scale, and $\eta$ is the residual gain.\cite{ref35}

In Eq.~(\ref{eq:conv}), $\odot$ indicates element-wise multiplication, and the first term modulates the frozen convolution output by a learnable magnitude offset normalized by the original weight norm. The second term adds a learnable residual path with a bottleneck ($A$, $B$) and depthwise mixing (DW), which allows for local adaptation of spatial location while minimizing the number of parameters used to produce it.\cite{ref35}

\paragraph{DiSCo spectral adapter:} We row-normalize the frozen weight by per-row norms $s_0$ to form $D=\frac{W}{s_0+\varepsilon}$, take a thin SVD $D\approx USV^\top$, and keep the top-$r$ subspaces ($U_r$, $V_r$).\cite{ref36} The adapted weight is as (\ref{eq:disco}):
\begin{equation}
W'=W+U_r\diag(c)V_r^\top+\diag(\delta_s)D.
\label{eq:disco}
\end{equation}
Here $c\in\R^r$ are learnable per-rank coefficients, $\delta_s\in\R^d$ are learnable per-output magnitudes, $r$ is the retained rank, and $\varepsilon>0$ ensures stability; optionally keep only the top-$K$ entries of $|c|$ at inference. DiSCo was integrated at 49 encoder sites, with no decoder or prompt encoder changes.

DiSCo offers a simple and direct method of separating direction and magnitude; the spectral term $U_r\diag(c)V_r^\top$ updates the directional component of $W$ within a low-dimensional subspace, while $\diag(\delta_s)D$ adjusts the per-output magnitude of the normalized weight. Thus, decomposition mirrors the motivation of weight-decomposed adaptation methods that decouple the scale of a weight vector from its relative direction to enable greater control of the weights in the context of parameter budgets.\cite{ref37}

For the optional Top-K inference constraint, we define an index set $\mathcal{K}$ of the $K$ largest components of $|c|$ and zero the remaining coefficients. One convenient representation is:
\begin{equation}
c_i^{(K)}=\begin{cases}c_i,&i\in\mathcal{K},\\0,&\text{otherwise},\end{cases}
\qquad W'=W+U_r\diag\bigl(c^{(K)}\bigr)V_r^\top+\diag(\delta_s)D.
\label{eq:topk}
\end{equation}
This retains only the most influential spectral components at inference, reducing effective adapter capacity while keeping the trainable parameter set unchanged.

\subsection{Evaluation Metrics}
We reported both segmentation quality metrics and compute-efficiency metrics to capture clinical utility (mask segmentation accuracy) and feasibility of deployment. Segmentation metrics were computed between the predicted tumor mask $P$ and the ground truth mask $G$, while compute metrics for segmentation were collected using a consistent inference protocol for all adapters.

\paragraph{Dice:} For binary masks, Dice is defined as:
\begin{equation}
\operatorname{Dice}(P,G)=\frac{2|P\cap G|}{|P|+|G|}=\frac{2\,\mathrm{TP}}{2\,\mathrm{TP}+\mathrm{FP}+\mathrm{FN}},
\label{eq:dice}
\end{equation}
where TP, FP, and FN denote true positives, false positives, and false negatives, respectively.\cite{ref28}

\paragraph{HD95:} We reported the 95th-percentile symmetric Hausdorff distance (HD95) between the predicted and ground truth mask boundaries to reduce sensitivity to outliers. Let $\partial P$ and $\partial G$ denote boundary point sets in physical units (mm). Using directed surface to surface distances, we computed the 95th percentile of directed distances in both directions and took the maximum:\cite{ref29}
\begin{equation}
\begin{split}
\operatorname{HD95}(P,G)=\max\Bigl(&\perc\bigl(\{\min_{g\in\partial G}\lVert p-g\rVert_2:p\in\partial P\}\bigr),\\
&\perc\bigl(\{\min_{p\in\partial P}\lVert g-p\rVert_2:g\in\partial G\}\bigr)\Bigr).
\end{split}
\label{eq:hd95}
\end{equation}

\paragraph{Compute metrics:} We reported (i) trainable parameters (M), (ii) inference latency (ms/image), (iii) throughput (images/s), (iv) peak GPU memory (GB), and (v) FLOPs per image (TFLOPs/img; lower is better). Peak memory was measured as the maximum GPU memory allocated during inference using the framework's built-in memory tracker.\cite{ref32} We also reported Dice per million trainable parameters (Dice/M) as a parameter-efficiency indicator.

\section{Results and Discussion}
\begin{figure}[tb]
\centering
\includegraphics[width=0.78\linewidth]{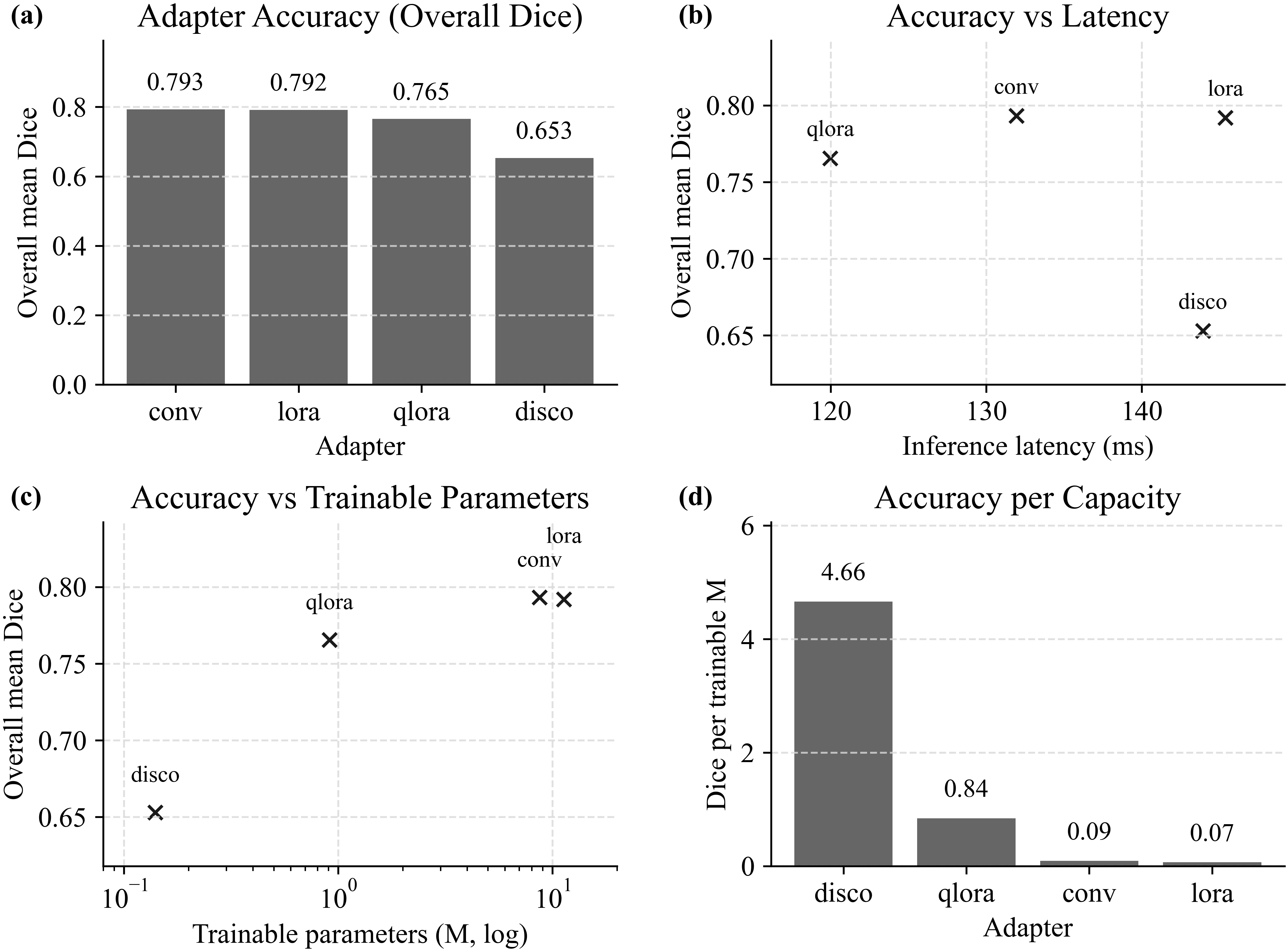}
\caption{Adapter accuracy--efficiency summary. (a) Mean Dice: Conv-Adapter/LoRA highest; QLoRA lower; DiSCo lowest. (b) Accuracy vs. latency: QLoRA fastest; DiSCo highlighted. (c) Accuracy vs. trainable parameters (log): Conv-Adapter/LoRA at higher capacity; QLoRA cuts parameters with moderate Dice drop; DiSCo in the extreme low-parameter regime. (d) Dice per million trainable parameters: DiSCo shows the best capacity efficiency.}
\label{fig:results}
\end{figure}

We evaluated four adapter strategies, Conv-Adapter, LoRA, QLoRA, and the proposed DiSCo, on all tumors within each patient under multiple prompt types (no prompt, single-point, multi-point, and box prompts at target IoU=0.50/0.75). We define overall Dice and HD95 as the patient-level mean averaged across these five prompt regimes; the single-point prompt achieved the best Dice performance among them. Figure~\ref{fig:results} summarizes the accuracy--efficiency landscape in a $2\times2$ panel: (a) overall accuracy (mean Dice), (b) accuracy vs. latency, (c) accuracy vs. trainable parameters (log scale), and (d) accuracy per capacity (Dice per trainable million parameters). DiSCo optimized capacity usage: with only 0.14 M trainable parameters it delivered the highest Dice per trainable million parameters (Fig.~\ref{fig:results}, panel (d)) and 6.95 img/s throughput; despite a lower Dice (0.6530), it occupied a favorable capacity--efficiency point under strict parameter budgets or small accelerators. For overall accuracy, Conv-Adapter is highest (0.7932 mean Dice), with LoRA essentially on par (0.7919; difference 0.0013), followed by QLoRA (0.7655) and DiSCo (0.6530). On the clinically common single-point prompt, the ordering is consistent: Conv-Adapter (0.7953), LoRA (0.7918), QLoRA (0.7680), DiSCo (0.6983) (Fig.~\ref{fig:results}, panel (a)). From a statistical perspective, the mean Dice gap between Conv-Adapter and LoRA is very small---0.0013 over all prompts and 0.0035 on the single-point prompt (Table~\ref{tab:performance}), corresponding to relative improvements of only 0.16\% and 0.44\%, respectively, whereas the gaps from Conv-Adapter to QLoRA and DiSCo are an order of magnitude larger (0.0277 and 0.1402 Dice, i.e., about 3.5\% and 17.7\% relative). Together with the HD95 values (LoRA only 0.27 mm better than Conv-Adapter but 4.67--17.79 mm better than QLoRA/DiSCo), this supports treating Conv-Adapter and LoRA as statistically comparable in accuracy and reserving clear superiority claims for the high-capacity (Conv-Adapter/LoRA) versus low-capacity (QLoRA/DiSCo) adapter families. Efficiency-wise, QLoRA shows the best runtime profile with the lowest latency (119.97 ms) and peak memory (4.94 GB), uses only 0.91 M trainable parameters, and has the lowest FLOPs per image (0.0020 TFLOPs/img); LoRA achieves high Dice but requires 11.31 M trainable parameters and has the highest FLOPs per image (0.9000 TFLOPs/img), while Conv-Adapter is intermediate (0.1860 TFLOPs/img) (panels (b)--(d)). In summary: for maximum accuracy choose Conv-Adapter or LoRA, whose mean Dice scores are statistically very close; for overall efficiency (latency/memory/parameters) choose QLoRA with a modest Dice drop; when parameter count or on-device storage is the hard bottleneck, DiSCo offers the strongest accuracy-per-parameter.

\begin{table}[tb]
\centering
\caption{Combined performance and resource summary on adapters. Dice and HD95 are patient-level means averaged across all prompt regimes; Single-pt reports the single-point regime. Resource metrics include inference latency, peak memory, trainable parameters, and FLOPs per image (TFLOPs/img); Dice/M reports Dice per million trainable parameters.}
\label{tab:performance}
\setlength{\tabcolsep}{3pt}
\resizebox{\linewidth}{!}{%
\begin{tabular}{lrrrrrrrr}
\toprule
Adapter & Dice $\uparrow$ & Single-pt $\uparrow$ & HD95 (mm) $\downarrow$ & Lat. (ms) $\downarrow$ & Mem (GB) $\downarrow$ & Train. (M) $\downarrow$ & FLOPs (TFLOPs/img) $\downarrow$ & Dice/M $\uparrow$\\
\midrule
DiSCo & 0.6530 & 0.6983 & 49.53 & 143.93 & 5.47 & \textbf{0.14} & 0.2807 & \textbf{4.657}\\
Conv-Adapter & \textbf{0.7932} & \textbf{0.7953} & 32.01 & 131.95 & 5.34 & 8.72 & 0.1860 & 0.091\\
LoRA & 0.7919 & 0.7918 & \textbf{31.74} & 145.38 & 6.09 & 11.31 & 0.9000 & 0.070\\
QLoRA & 0.7655 & 0.7680 & 36.41 & \textbf{119.97} & \textbf{4.94} & 0.91 & \textbf{0.0020} & 0.842\\
\bottomrule
\end{tabular}}
\end{table}

These results are consistent with the expected capacity hierarchy of PEFT techniques, where higher capacity adapters (Conv-Adapter or LoRA) offer the strongest accuracy, while aggressively parameter constrained ones (QLoRA or DiSCo) trade some accuracy for improved deployability. In practical use cases, single-point prompting is an attractive option because it strikes a balance between user workload and performance, while box prompting can provide a means to quickly refine the volumetric extent of a lesion based on prior knowledge of its approximate location (e.g., by selecting an ROI by a radiologist), which is important for expediting workflow.

A key clinical implication is that PEFT has enabled institutions to specialize foundation models to accommodate differences in scanner or protocol characteristics without incurring the costs associated with completely retraining a model through fine-tuning. This is of particular importance to institutions that require rapid adjustments to meet local imaging needs or hardware limitations while preserving the ability to utilize a uniform promptable interface.\cite{ref17,ref24,ref25}

A limitation of this research is its reliance on 2D slice export (which does not directly model 3D context) and it was evaluated with a single in-house patient cohort. Future studies will evaluate the robustness of PEFTs (in regards to more diverse acquisition) across multiple institutions and with 3D-aware adaptation techniques while preserving their efficiency.

\section{Conclusion}
Adapterized PEFT of SAM\cite{ref17} enables accurate, deployable CT liver tumor segmentation while training far fewer parameters and supporting point or box prompts. We freeze the SAM backbone and train only lightweight adapters (Conv-Adapter, LoRA, QLoRA) or the proposed DiSCo, which performs SVD on row-normalized weights to form spectral bases and learns compact spectral coefficients with per-output offsets and a scalar gain, with an optional inference-time Top-K constraint that retains only the largest-magnitude spectral components. As summarized in Fig.~\ref{fig:results} and Table~\ref{tab:performance}, in CRLM Conv-Adapter and LoRA lead on accuracy; QLoRA offers the best accuracy--efficiency trade-off (low memory and latency); and DiSCo delivers the highest parameter efficiency for ultralimited budgets. Overall, this work supports rapid scanner-specific tuning, enabling practical preoperative volumetric workflows, surgical planning, and radioembolization.

\section{Acknowledgments}
This work was funded by the National Institutes of Health (NIH) and the National Cancer Institute (NCI) through grants R01CA233888 and U01CA238444.

% Bibliography transcribed from the source manuscript; reference order preserved.

\end{document}